\documentclass[10pt]{article}

\usepackage{mathrsfs}
\usepackage{epic, eepic}
\usepackage{subfigure}
\usepackage{amsfonts, amssymb, amsmath, graphicx, epsfig, lscape, capt-of}
\usepackage{url}
\usepackage{ifpdf}
\usepackage[boxed]{algorithm2e}
\usepackage{array}
\usepackage{booktabs}
\usepackage{multirow}
\usepackage{xcolor}
\usepackage{adjustbox}

\def\inst#1{$^{#1}$}

\makeatletter

\begin{document}%


\title{Words ranking and Hirsch index for identifying the core of the hapaxes in political texts}

\author{%
Valerio Ficcadenti\inst{1,}\thanks{Corresponding author.}\and Roy Cerqueti\inst{1,2}\and
Marcel Ausloos\inst{3,4} \and
Gurjeet Dhesi\inst{1} }
\date{}

\maketitle

\begin{center}
{\footnotesize

\vspace{0.3cm} \inst{1} School of Business\\
London South Bank University\\
London, SE1 0AA, UK\\
\texttt{ficcadv2@lsbu.ac.uk; cerquetr@lsbu.ac.uk; dhesig@lsbu.ac.uk}\\
\vspace{0.3cm} \inst{2} Department of Social and Economic Sciences \\
Sapienza University of Rome \\
Rome,  I-00185, Italy\\
\texttt{roy.cerqueti@uniroma1.it}\\
\vspace{0.3cm} \inst{3} School of Business \\ University of
Leicester \\ Brookfield, Leicester, LE2 1RQ, UK\\
\texttt{ma683@le.ac.uk}\\
\vspace{0.3cm} \inst{4} Department of Statistics and Econometrics \\
Bucharest University of Economic Studies\\
Bucharest, Romania}

\end{center}

\begin{abstract}
This paper deals with a quantitative analysis of the content of official political speeches. 
We study a set of about one thousand talks pronounced by the US Presidents,
ranging from Washington to Trump. In particular, we search for the
relevance of the rare words, i.e. those said only once in each
speech -- the so-called \emph{hapaxes}. We implement a rank-size
procedure of Zipf-Mandelbrot type for discussing the hapaxes'
frequencies regularity over the overall set of speeches. Starting
from the obtained rank-size law, we define and detect the \emph{core
of the hapaxes} set by means of a procedure based on an Hirsch index variant. We discuss the resulting list of words in the light of the overall US
Presidents' speeches. We further show that this core of hapaxes
itself can be well fitted through a Zipf-Mandelbrot law and that
contains elements producing deviations at the low ranks between
scatter plots and fitted curve -- the so-called \emph{king} and
\emph{vice-roy effect}.
Some socio-political insights are derived from the obtained findings about the US Presidents messages.
\end{abstract}

\section{Introduction}

Scientific debate has recently grown on text analysis and data
mining because of the relevance of the information taken from texts and
for the need of a systematic quantitative analysis of them. For example, it is worth mentioning \cite{Lambiotte2007}, where the authors study the regularities of words occurred in blogs and \cite{jiang}, where the authors propose a model for assessing borrowers' defaults on loans by analyzing texts on the available descriptions of such loans.
In \cite{chan} the authors pay peculiar attention to the exploration of the financial texts for their relevant informative content.
In the same context, in \cite{yuan} there is a discussion of the determinants of the crowdfunding successes by following a text analysis approach.
\newline
Nowadays, politicians use social networks to inform their voters, therefore they calibrate the messages on the bases of objectives to be addressed (see \cite{Yoon2016} for the case of Seoul mayoral election). The official speeches of the US Presidents are, of course, carefully written. Each single locution or term is evaluated, in order to guess what the impact will be on the audience and in the entire
socio-economic environment.
\newline
This paper begins with the above premise and applies it to the analysis of some relevant aspects of a large set of Presidents' speeches. In doing so, we are in line with the studies of the communication of US Presidents (and also candidates to presidency or Presidents' media importance \cite{Park2016, Takikawa2019}) and its socio-economic relevance (see e.g. \cite{cochran, pollaw}).
\newline
Our target is to assess the presence of regularities in the
frequencies of the hapaxes and explore the existence of a qualified
set of words pronounced only once in a large number of speeches. Such regularities are able to outline a scheme for supporting decisions in communication task. It is important to point out that the methodology here designed can be used in any type of corpora whose hapaxes follow a rank--size power law behaviour.
With this aim, we study the collection of the hapaxes in each speech.
\newline
But why are hapaxes so relevant in the official context of the US
Presidents' speeches, and why is it worthy to explore them? Hapaxes
might represent subliminal-like messages that the US Presidents
deliver to the audience. 
Furthermore, hapaxes play a fundamental role in identifying the
richness of a speech. Thus, the use of hapaxes represents an
instrument for informing the audience that the President has a high
level of culture and education, hence giving him credit with
electors and relevant institutional figures. Under this perspective,
it is important to mention one of the most commonly used
morphological productivity measure, i.e. (see \cite{Bauer2001}):
$$
P=\frac{number\, of hapaxes\, in\, a\, text}{length\, of\, the\,
text}.
$$
It is also important to stress that the most part of the hapaxes is associated to some
peculiar characteristics, which lead to crucial research questions and conjectures (see the long list of future research outlined in the last section).
\newline
The dataset here considered comes from the rough data contained in
the Miller Center, which is a research institute affiliated to the
University of Virginia (see \url{http://millercenter.org}). A set of
about 1000 US Presidents' speeches has been downloaded from such a
website, from the \textit{Inaugural address} of George Washington
(1789) to the Donald Trump's speech \textit{Address to Joint Session
of Congress} (2017). Presidents' contributions like \cite{Obama2016} are excluded; indeed they do not appear into the Miller Center database and they are not framed as speeches, but as scientific papers.
\newline
The treatment phase of the speeches -- whose details will be
presented in Section \ref{data} -- has been attained through data mining techniques
(for a survey on text mining, see e.g. \cite{bookcambridge, book2}).
\newline
The study proceeds in three sequential directions (steps).
\newline
Firstly, a rank-size relation has been assessed over the set of the
hapaxes, where \emph{size} is measured through the frequency of the
words in the entire set of speeches. For an excellent review of the
empirical settings where power laws is a valid device for
representing related phenomena and some theoretical explanations of
such a way to fit data, we refer to \cite{newman}. In agreement with
other linguistic studies
\cite{montemurro2001beyond,piantadosi2014zipf,
ausloos2008equilibrium, ausloos2010punctuation,ausloos2016quantifying, rovenchak}, we have tested the
validity of the Zipf-Mandelbrot law in properly fitting the data by
implementing a best-fit optimization procedure
\cite{zipf1935psycho,zipf1949human,mandelbrot}. In this preliminary
phase, we have found statistical compliance of the
considered dataset with Zipf-Mandelbrot law, even in presence of
(quite) negligible deviations at low ranks (see the fourth step
below for a comment on this).
\newline
In the second step, we have used the obtained calibrated curve to
identify the core of the hapaxes by using the indicator proposed in \cite{ausloos2013}, in the science measurement context of the scientists'
coauthors. Such an indicator is a
replication of the $H$-index -- where $H$ stands for Hirsch, who
invented it in \cite{hirsch} -- used to evaluate scientific
research (see \cite{guns2009} for a more detailed description and \cite{Schreiber2010} for a comparison with different variants). In this context, the core of the
hapaxes is the set with cardinality $\bar{H} \in \mathbb{N}$ which
contains the maximum number of hapaxes whose frequency is at least
$\bar{H}$. In this respect, the ratio between the area of the core
and the one of the entire set of hapaxes -- computed with respect to
the best-fit curve of the rank-size law -- is a percentage measure
of the most relevant hapaxes in the overall history of the US
Presidents' speeches.
\newline
The third step consists in the exploration of the core and of its
properties. We here show that the core is a set whose hapaxes have
ranked frequencies again satisfying a Zipf-Mandelbrot law.
Furthermore, as already pre-announced above, in the present rank-size
analysis of the overall set of the hapaxes, we have found small
deviations at low ranks.
This means that the best fit curve does not represent "perfectly" the scatter plot of the
low-ranked hapaxes. The reason for this stands in the outlier-type
behaviour of a group of hapaxes which are contained in several
speeches. We here guess that such outliers are the hapaxes in the
core and redo the best fit procedure by removing the core from the
overall original sample. Results confirm the improvement of the fit.
According to \cite{ls}, the token at rank equal to 1 is the
so-called \emph{king} whilst the others are the \emph{vice-roys},
and in this case there is a \textit{king plus vice-roy effect}. For
a further example of this effect, refer to \cite{ACphysA}. An
interpretation of such an effect will be also presented in the Section devoted to discuss the results.
\newline
To sum up, the contributions of this paper relate to the well known rank-size behaviors of words frequencies in textual data; in particular, we address the peculiar case of the hapaxes distribution in corpora. We propose an innovative method to determine relevant hapaxes; the approaches used and the findings have a comparable impact to that of \cite{Morente2018}, where the authors propose an innovative approach to analyze opinions in social network and to determine decision making processes on the basis of users' sentiments.
\newline
We point out the presence of a link between this paper and \cite{Ficcadenti2019}. Indeed, analogously to the quoted paper, we here move from the discourses retrieved from the Miller Center website, and opportunely treat them to extract individual constitutive words with the related frequencies for each of them.
\newline
Differently, we here focus only on the words said only once in any discourse. In so doing, we are radically different from \cite{Ficcadenti2019} in three main directions: firstly, about the data, the object of the analysis of the present paper is a subset of the source database used in the quoted paper. Here we analyse just the words with frequency one in each speech, while in \cite{Ficcadenti2019} all the words are considered. Consequently, the data processing phase is a refinement of the one employed for collecting and treating all the pronounced words;
secondly, the scientific ground of \cite{Ficcadenti2019} lies in the aim of understanding and comparing the regularities of the structures of the individual speeches in terms of words frequencies, while the ground of the present report lies in understanding the way in which the same words -- i.e., those used only once in each speech -- have been historically employed by the US Presidents; thirdly, sometimes the Presidents want to communicate messages with a certain degree of discretion, targeting a specific subset of the audience. In this respect, the hapaxes analysis maps the unique reference to less central topics treated in a speech. Their investigation, especially on such a dataset, gives hints about the informative content and the target of the presidential communications. In so doing, the results provides a wide number of research questions, to be addressed in future studies.
\newline
The rest of the paper is organized as follows. Section
\ref{Motivations} contains a wide discussion on the reference literature. In Section \ref{data} there is a presentation of the procedure employed to collect the data. Section \ref{method} is devoted to the
illustration of the methodology used for the analysis. Section
\ref{result} presents the results and related comments.
The last section offers some conclusive remarks; moreover, it contains also a specific focus on future research themes.


\section{Literature review}
\label{Motivations} This section contains a brief review of the literature, to support 
the research and scientifically motivate the worthiness of the
proposed study. We start with a discussion of the hapaxes; then, we present a critical view of the employed methodology.  

\vspace {0.3cm}

\noindent The exploration of the hapaxes goes much further than usual text
content or structure analyses; hapaxes have a special meaning
(see e.g. \cite{auslooshapax, anal1,drozdz,ausloos2012b, denghapax,
metinhapax}). Some remarkable examples are worth mentioning.
\newline
In the overall work of Giacomo Leopardi, the word
\emph{ultrafilosofia} has been used only once for the
contextualization of the philosophical system of the author.
However, the authoritative Encyclopedia Treccani refers to
\emph{ultrafilosofia} to describe Leopardi's thought. In the
related entry, \emph{ultrafilosofia} is no longer a hapax, but
appears 9 times \cite{polizzi}.
\newline
\emph{Mnemosynus} is a hapax for the Latin language. In fact, it
appears in the entire collection of available writings in Latin only
once, in Catullo's Carmina. This term points to the mythological
figure of the goddess of memory. Such a hapax has been not neglected
in subsequent modifications and contaminations of the linguistic
evolution, and \emph{mnemonic} comes evidently from
\emph{Mnemosynus}.
\newline
It is important to mention also the relevance of the hapaxes in the
holy books of Bible and Quoran \cite{hapaxbible, hapaxquoran}, which
contain speeches attributed to one or several authors.
\newline
Thus, it is not unexpected that some authors have dealt with the
analysis of the hapaxes.
\newline
In \cite{Joandi2012}, the authors state that the presence of hapaxes in
a text can be used to determine the language productivity of terms,
so the language inflection (see also
\cite{Bauer2001} for related material). Therefore, studying the
hapaxes of corpus from a common source along the years allows to
capture neologisms (e.g. \cite{Bauer1983}). Consequently, it is
useful to interpret changes into a community of people speaking
a common language as presented in \cite{Lipka2010}.
The study of the hapaxes across different documents is important in authorship
attribution as well (see \cite{Iqbal2011}, where a wide description of the field is reported). For example, in \cite{Holmes1994}, the author states that it is
possible to test the tendency of an author to choose between a word
used previously or utilizing a new word instead.
 In \cite{Smith2002} is outlined the importance of the hapaxes within a corpus to extract
information about the writers stylistic changes.
\newline
To highlight the potentiality of the hapaxes analysis in spreading political messages and in supporting decisions about messages' structures, it is worthy to mention the different Figures of Speeches (for a review of them, see \cite{Soules2015}). Indeed,
there are specific figures like the \textit{climax} used to increase
the importance of a concept by intensifying the usage of certain
words. On the other hand, by negation, if the message has to be
hidden rather than stated, mentioning a word just once is enough to
relay the communication. The analysis of hapaxes has never been
employed in this respect but the hapaxes legomena are often involved
in political communications analysis, e.g. see \cite{Savoy2010} for
the specific case of US President candidates John McCain and Barack
Obama.
\newline
In general, some political speeches have hidden embedded information
about the future, see for example `Address at Moscow State
University` stated in May 31, 1988, by President Ronald Reagan, less
than one year before the fall of the Berlin Wall, where the words
``freedom'' is repeated more than 20 times. Is it significant or
related to the fall of the Berlin Wall occurred shortly thereafter?
\newline
The US President represents one of the large powers in the world, so
it is easy to guess that they have a large competitive advantage in
gaining information with respect to their audience (President Nixon
is a clamorous and ironic example of that!). The US Presidents have
classified information about potential wars and treats for the
country, so they might aim at preparing the public opinion for
potential conflict by evoking specific sentiments. For example this
could be the case of US before the World War II, when President
Roosevelt started to consider an involvement of US into the Britain
- French and German conflict (see the Fireside Chat 15 and 16 of
President Roosevelt, the former titled `On National Defense' and the
latter `On the Arsenal of Democracy').
\newline
Therefore, in the very special context of political communication,
we notice that the US Presidents' speeches are written in a so
precise and careful way that the presence of a word pronounced only
once cannot be seen as accidental. However, the occurrence of a hapax
in one speech does not tell credible stories \textit{per se}.
Despite this evidence, the presence of regularities in the selection
of a given hapax over a wide number of speeches points out to a
common behaviour of the Presidents, and merits attention. This is the
scientific ground of our study.

\vspace {0.3cm}
\noindent We now provide a critical discussion of the methodological devices used in the paper, along with the most relevant literature contributions. 
\newline
The employment of regression techniques of rank-size type in the
text analysis finds also support in the literature. For the special
case of the exploration of hapaxes within a rank-size framework, we
refer to \cite{Mohammadi2016, MorinHazem2015}, where the authors
employ Zipf's law and point out that the relevant analyses of
hapaxes, helps in the exploration of parallel documents.
\newline
In this respect, it is important to note that rank-size analysis
allows to derive a panoramic view of a unified system generated by
granular data -- the frequencies of the hapaxes.
\newline
We also point out that the definition of the core through the Hirsch
$H$-index exploits the meaningfulness of such an index. In
particular, the $H$-index is able to synthesize the overall number
of hapaxes and the frequencies with which they appear in the set of
the speeches in a unique entity. To fully understand this point we
refer to the familiar use of $H$ in the research evaluation context,
where $H$ gives a clear idea on the overall productivity of a
scientist and on the impact -- in terms of citations -- of her/his
production on the scientific community, i.e. the so-called "core of publications".
In this line, one can extend the Hirsch-index idea to other cases, like to define the "core of coauthors" of a researcher (see \cite{ausloos2013,auslooshapax}). Generalizing the idea, we define the core of hapaxes in the investigated texts.  Let us briefly recall the logical aspects.
 \newline
Indeed, it was found out that   a  Zipf-like law
 \begin{equation}   \label{eqZipf}  J \propto 1/r ,
  \end{equation}
  exists,  between
the number ($J$) of  joint publications (NJP) of a scientist, called for short "principal investigator" (PI) with her/his coauthor(s)  (CAs); $r $ =1,...   is an integer allowing some hierarchical ranking of  the CAs; $r=1$ being the most prolific co-author of the PI.    Yet, it   was  observed that  a hyperbolic  (scaling) law   is more appropriate, i.e.,
 \begin{equation}    \label{eq1}
 J =  J_0/r^{\alpha} ,
  \end{equation}
 with $\alpha\neq1$, usually  such that  $\alpha\le1$, and often decreases with the number of CAs or with the number of joint publications,  e.g. when the number of CAs  and when $J$ are   "not  large". $J_0$  is a fit parameter, i.e. there is no meaning to $r=0$.
 \newline
 We can follow such a line of thought for the hapaxes, being ranked, and noticing those below a given threshold. As   the $H$-index \cite{hirsch,hirsch10,rousseau06}   "defines"  the {\it core of papers of an author} from the relationship between the number of citations $n_c$ and the corresponding rank $r$ of a paper, through a trivial threshold,  i.e. if $n_c$ $\ge r_c$, then $r_c\equiv h$,  thus
  one is allowed  to define  the {\it core of  hapaxes} through  a  threshold  \cite{ausloos2013,auslooshapax},
 called  the  $m_a$-index, for short,
   \begin{equation} \label{eq2}
 m_a \; \equiv \; r,   \;  \;  \;  \;  as  \;long\;  as\;\; \; \; r\; \le \; J.
  \end{equation}
Technically, one could thus measure the relevant strength of a hapax, whence measure  some impact of such a word on speech intention, as in research collaborations  \cite{LeeBozeman}. This is exactly the methodological perspective adopted in the current paper.

\section{Data}
\label{data}
\begin{figure}
\includegraphics[width=\textwidth]{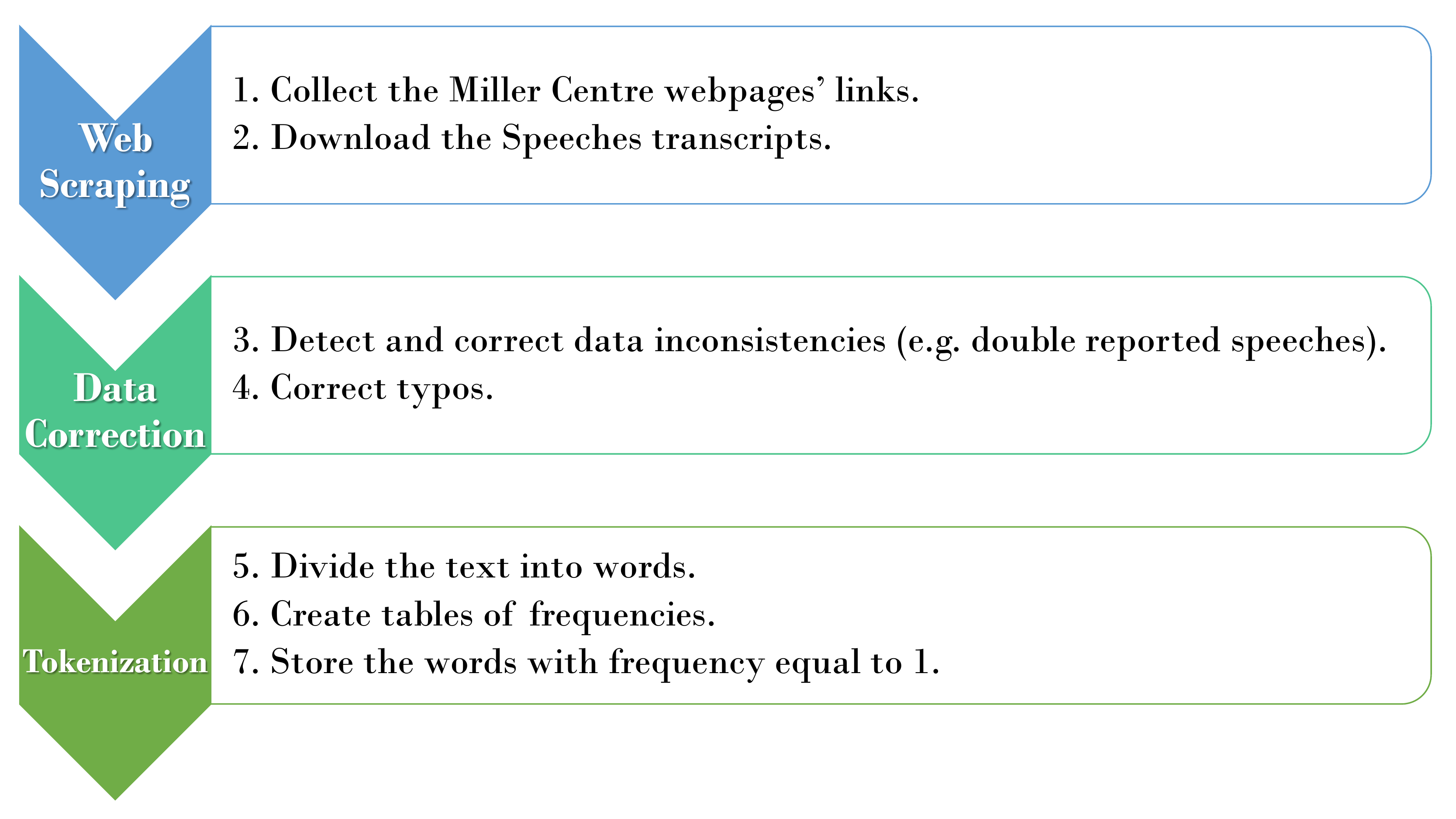}
\caption{Summary of the necessary phases to collect and store the data used in this paper.}\label{diagram}
\end{figure}

\begin{figure}[ht]
\includegraphics[width=\textwidth]{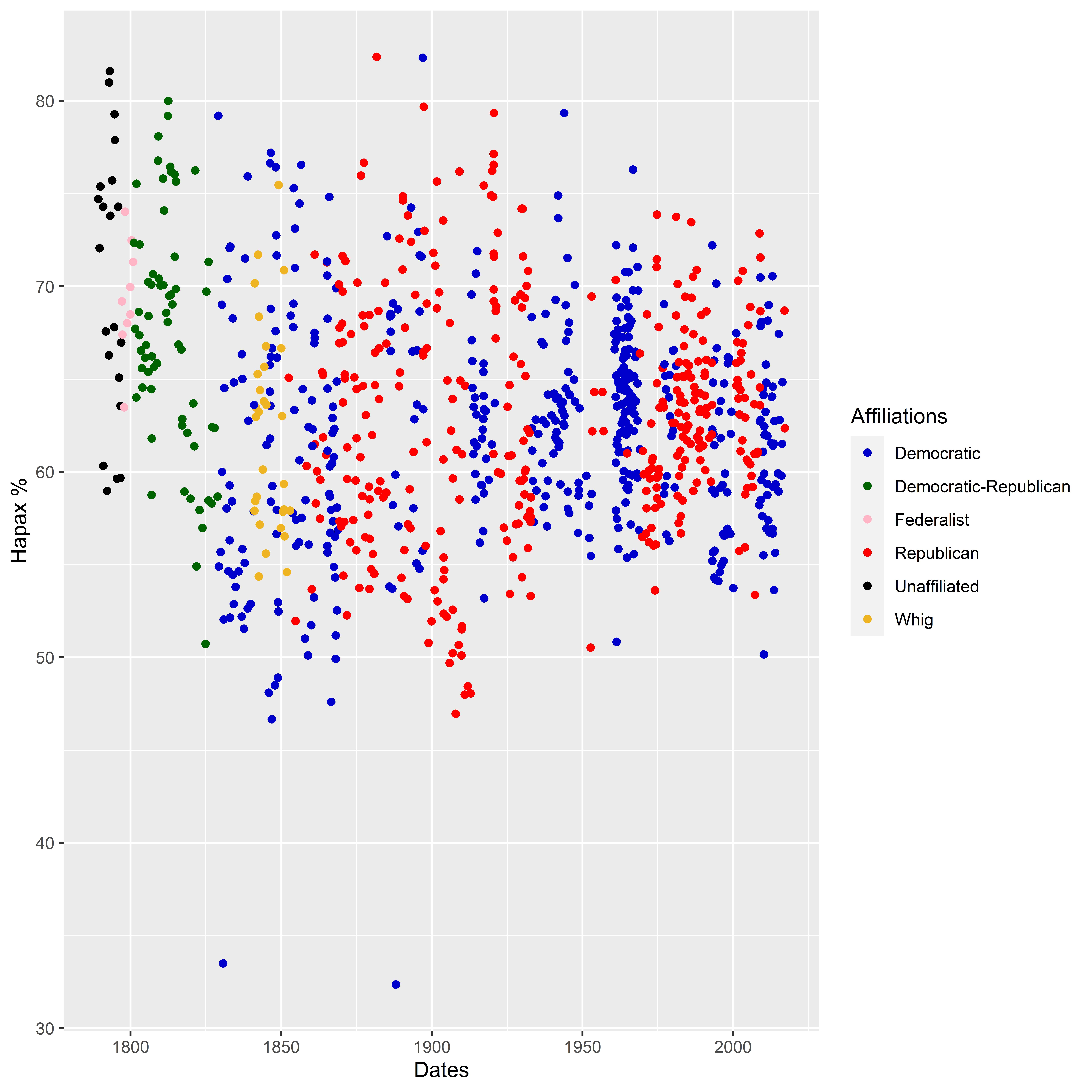}
\caption{Percentage of hapaxes per speech ranked in chronological order. The colours represent the speakers' affilations; they are reported for the sake of clarity.}\label{hapaxperc}
\end{figure}

This section is dedicated to describing the data collection process. It is very close to the one presented in \cite{Ficcadenti2019, Cinelli2019}, hence we report here just the phases implemented to get the hapaxes and to improve the original procedures. The key process' steps are summarized in Figure \ref{diagram}.
\newline
{As a premise, a methodological remark is in order. Of course, the
adopted strategy for analyzing the speeches includes also personal
visual inspection. However, the general procedure for the pre-processing and web scraping phases are stepwise listed.
In this way, it is possible to replicate the study.}
\newline
{Data have been retrieved from the Miller Center, which is a research center affiliated to the University of Virginia (see \url{http://millercenter.org}).}
The Miller Center is still active and adds
continuously speeches. Thus, the codification of the problems
fosters future updates of the analysis of the hapaxes.
\newline
{We point out that a conservative principle drives our actions, so that the original texts are modified as little as possible. In so doing, we go in the direction of maintaining the highest level of similarity of the final dataset with the original speeches. At the
same time, all the errors into the transcripts -- including those of minor nature -- are taken into full consideration. This required additional efforts with respect to \cite{Cinelli2019}}, in fact, we have checked the text by using the \textit{``hunspell''} R package \cite{hunspell}, with the English dictionary. We extract and correct all the spelling errors within each talk. Nevertheless, there is still a list of 7716 exceptional tokens that are not found in the English dictionary. They are considered as potential typos that have to be investigated one by one. Many of them are not exactly errors, but they are exceptions invoked by the speakers for satisfying specific rhetoric needs or past ways of speaking not comprised into the US Hunspell English dictionary. Some examples are given by the usage of peculiar non-English personal names like ``Bernardino'', terms from Spanish or French like: ``intendencia'' or ``arrete'' and terms that were differently spelt in the past like: ``regrassing'' or ``tofore''. Even if these types of exceptions enter into the list of potential typos, they cannot be considered
errors to be modified because we assume that the Presidents have pronounced them into a specific context that required such uses. Therefore, to make the distinction between the exceptions just described and the flaws that have to be modified, we have looked at all the lists of potential typos into the \textit{Cambridge
Dictionary}, the \textit{Oxford Dictionary} or \textit{Wiktionary}. In this way, it is possible to identify ancient English uses no longer in vogue, foreign words or common language flaws in accord to
the proper linguistics uses. When there are not any straight suggestions, one has to extract the entire phrase that contains the ambiguous terms from the respective transcript (the whole statements are easily captured by looking for them into the corpora). Then, thanks to the exact search of \textit{Google}, it is possible to check if the ambiguous words are reported into other speeches' transcripts sources by examining the results of the research. If there are other references for the same phrases with the terms corrected, we adopt the most logical usage by interpreting the meanings of the findings\footnote{Examples of this type are given by the typos ``questionin'', ``lawon'' and ``adispute'', that come from the Inaugural Address stated by Rutherford B. Hayes, March 5, 1877. The bugs are in the following phrase: \textit{``The fact that two great
political parties have in this way settled adispute in regard to which good men differ as to the facts and the lawno less than as to the proper course to be pursued in solving the questionin
controversy is an occasion for general rejoicing.''} One can intuitively guess that the corrections of the wrong terms are
``question in'', ``law on'' and ``a dispute'' but for acting systematically and for being coherent with the method adopted, the exact search is run. In the context of the example, the research returns many other sources where the words' correct forms are adopted as expected.}. The potential flaws with the highest degree of
uncertainty are adjusted or not on the bases of the majority criterion. It means that one uses Google's exact search and then adopts the most common phrases within the first ten results hereby found. Thanks to this correction process, 3851 improvements are applied. Consequently, the remaining locutions cannot be considered wrong. This step is the only one requiring human judgment; therefore, it is the only reason to define the nature of the procedure as \textit{semi-automatic}.
\newline
{In this light, one has to read also the choice of not considering the different variations of the same words -- like the different forms of the same verb, singular and plural, etc. -- as different words. We have a twofold explanation for such a choice. 
\newline
First, we point out that the analysis is implemented on the tokens in the speeches, with special attention to the tokens said only once. We recall that our study starts from the crucial premise that speeches are carefully written. In this respect, any word is likely selected in light of a communication strategy. In this context, implementing a sort of regularization process for all the tokens -- for example, putting all the verbs in infinite form -- might provide an arbitrary misinterpretation of the communication target of the speech. Substantially, we are not fully authorized to consider ``give" and ``given" as the same word. The President may decide to conveniently select the present tense or the past tense for the same verb in according to the contingent needs. Second, such a regularization process would lead to an additional source of biases and inconsistencies when handling ancient English -- as the discussion above clearly illustrates.}
\newline
The whole resulting list of words pronounced only once per speech is made of 31074 tokens, with frequencies that range in $[1, 250]$. It means that there is a term used just once in 250 speeches and another that appear one single time only. The principal statistical indicators can be found in Table \ref{stat}, column (a), which allows having a view of the frequencies' collection properties. Figure \ref{hapaxperc} shows the percentage of hapaxes per speech along with the speaker's affiliation (this last information is there just to enrich the information carried out by the figure).

\begin{table}
\centering
\begin{tabular}{ccc}
\hline
\textbf{Statistical indicator} & \textbf{Whole corpus (a)} & \textbf{Core hapaxes (b)}\\ \hline
N. Words & 31074 &  182 \\
Mean ($\mu$) & 16.3850 & 199.6484 \\
Variance ($\sigma^2$) & 1034.2965 &   183.279 \\
Standard deviation ($\sigma$) & 32.1605 & 13.5381 \\
Skewness & 3.2451 &  1.1188 \\
Kurtosis & 11.5989  & 1.463165 \\
Median ($m$) & 3 & 197 \\
Max & 250 & 250 \\
Min & 1 & 183 \\
RMS & 36.0934  &   199.644 \\
Standard Error & 0.1824 & 1.0035 \\ \hline
$\mu/\sigma$ & 0.5095 & 14.7472 \\
$3(\mu-m)/\sigma$ & 1.2486 &  0.5869 \\
\hline
\end{tabular}
\caption{Main statistical indicators associated to (a) the whole list of hapaxes in the
dataset and (b) the set of the hapaxes belonging to the core.}\label{stat}
\end{table}

Despite the applied correction process, some minor typos are still reported into the hapaxes dataset. They do not exceed 2\%. This level of error is expected for a dataset made by speeches transcriptions and it is unavoidable also for linguistic reasons.
Indeed, we are not able to resolve some linguistic ambiguity of certain terms, whence inducing some inaccurate error bar. However, we used computer-based algorithms and human readings to reach an expectedly very low set of possible forgotten cases of misspellings. For example, the word ``unmunitoned" which appears in ``Address at the Celebration of the 150th Anniversary of George Washington Taking Command of the Continental Army, Cambridge, Mass"
stated by Calvin Coolidge on July 3rd, 1925, can be a misspelt word or not. We prefer to consider it as a typo contributing to the stated 2\% since we are not sure of its meaning; a linguistic investigation -- unmonitored, at the moment -- would be appropriate for cases like this. Furthermore, even if one decides to go through each residual terms, there remains the possibility of leaving errors due to natural human predisposition to commit an error in a manual control -- operational risk. The idea of using a coded routine for spell checking the ambiguous words is a side scientific product of the research; this is an improvement to \cite{Ficcadenti2019}. In this way, we make a procedure that defines what can be considered as typo into the context of this framework. Anyway, a visual inspection of the remaining terms allows concluding that the majority falls into the terms that occur just once into the hapaxes' list. Therefore, this reinforces the idea that the residual cases lead to a negligible effect on the analysis object of the present study.
\newline
Yet, as we will show in the section devoted to the methodology, the procedure to determine the
analyze the hapaxes and identify their ``core'' combines the H-index
(see \cite{hirsch}) with the Zipf rank-size law. In particular, the
core consists of a group of hapaxes which are the most frequent ones in
the US Presidents' speeches dataset. The definition of the core
guarantees that the possibly existing typos are excluded from it,
and can be found just in the tail of the list of the hapaxes (sorted
by frequency). Indeed such tokens, being typos, have very low
chances of occurring more than once (namely in more than one
speech), and appear only in the speech containing them.
\newline
Besides, even if the presence of typos in the tail of the hapaxes list could affect the estimation of the parameters of the Zipf-Mandelbrot law, a so small percentage of 2\% is statistically considered not to lead drastic changes which could be harmful to the analysis.

Before moving to describe the methodological devices used for the analysis, a remark on the worthiness of the dataset is needed.
\newline
The corpus we have used in this study is the same corpus that has been employed also in \cite{Cinelli2019, Ficcadenti2019}. To the best of our knowledge, it is one of the most complete that it is possible to find in the literature. To support this point, we mention studies in which the authors have analysed institutions' communications:

\begin{itemize}
\item In \cite{Savoy2010}, the authors have used 245 US Presidents' speeches.
\item \cite{rule2015lexical} is an outstanding study of the State of the Union discourses in US. The set of analyzed speeches is a sensible subset of our corpus.
\item In \cite{light2014words} the analysed data consists of the corpus of United States' presidential Inaugural Addresses from 1789--2009. Also in this case, the dataset is a sensible subset of our corpus.
\item The contribution \cite{schonhardt2012yes} focuses only on the Ronald Reagan's discourses.
\item In \cite{kahveci2016central}, the authors have used about 150 Central Banks communications. They analyzed different monetary institutions, but for each of them they have used about 150 communications.
\end{itemize}

It is also important to point out that it is frequent to find researches whose target or scientific analysis ground is given by a restricted number of communications delivered by institutions and political subjects. Unfortunately, the absence of digitalization of old texts sometimes prevents the employment of a complete corpus mainly for difficulties in having access to the original texts or for the lack of scanning systems and precise enough tools like the latest OCR (Optical Character Recognition) technologies. In the case treated in the present paper, we are free from this problem, since almost all the speeches delivered during the US history are available and accessible on a website. Furthermore, the procedure designed in this paper is particularly appropriate to explore corpora made by a non-enormous amount of texts.

\section{Methodology}
\label{method}

The hapaxes of the individual speeches have been merged together in
a unified list. Each word is associated to an integer, which assigns
to it the number of speeches in which such a word is a hapax. We
briefly call \emph{frequency} this number, so that the resulting
list is composed by a series of words with associated frequencies.
\newline
The resulting list of hapaxes contains 31074 words. The maximum
frequency is realized by the word \emph{sense}, which appears 250
times as a hapax in a President's speech. Moreover, there is a list
of 10088 tokens which are hapaxes only in one speech (thus, having
unitary frequency).
\newline
For the details of the construction of the dataset, see the Section \ref{data} of this paper.
\newline
Hapax words are ranked in decreasing order, according to their
frequencies. In this respect, the ''size'' of a word is its
frequency. In the rank-size analysis, we will denote size and rank
by $s$ and $r$, respectively.
\newline
The Zipf-Mandelbrot law is used for best fit search, according to the following
rule:
\begin{equation}
\label{zipf} s=f(r)=\frac{\alpha}{(\beta+r)^\gamma},
\end{equation}
where $\alpha, \beta, \gamma$ are parameters to be calibrated for
fitting the sample under investigation.
\newline
As we will see, there is a very good compliance of the considered
data with the Zipf-Mandelbrot law (see rows (a) of Table \ref{Table1}, and Figure
\ref{Fig1} in Section \ref{result}). Such a property can be used to
define the measure of the core of the hapaxes.
\newline
%
%
%
%
%
%
In fact, the core of the hapaxes is defined through the $H$ index,
in a similar way in which it has been introduced by \cite{hirsch} to
evaluate scientific research. Specifically, such an index is
$\bar{H}$ when $\bar{H}$ is the maximum number of words whose
frequency is at least $\bar{H}$. The resulting set of $\bar{H}$
words is the core of the hapaxes.
\newline
In practical terms, the $H$-index provides a formal identification of the contrast between the most frequent hapaxes and the less frequent ones. Such an indicator provides also a relevant information on the overall distribution of the hapaxes -- thus, also on the way in which US Presidents have historically decided to pronounce some specific words only once in a speech -- since it is strongly dependent on how ranked data are positioned in terms of their sizes. In this respect, it is worth pointing out that a low (high) value of $H$ stands for a large (small) distance between hapaxes which are consecutive at high ranks. This outcomes suggests that the value of $H$ allows to explain if US Presidents focus obsessively on a few specific words to be pronounced only once (case of low value of $\bar{H}$) or, conversely, such an imitative behaviour does not take place.
\newline
By employing $\bar{H}$ and the best-fit curve defined in
(\ref{zipf}), with parameters in Table \ref{Table1}, block (a) -- justifications for choosing such estimated values will be detailed later -- we are able
to provide an absolute and relative measure of the core of the
hapaxes. We denote such measures as $\cal{M}_A$ and $\cal{M}_R$,
respectively. They are defined as the area of the region below the
curve in (\ref{zipf}) delimited by $r=1$ and $r=\bar{H}$ and as the
ratio between such area and the area of the overall region, from
$r=1$ to $r=31074$, respectively. Specifically, the absolute measure
of the core of the hapaxes is

\begin{equation}
\label{MA}
\mathcal{M}_A=\int_{1}^{\bar{H}}\frac{\hat{\alpha}}{(\hat{\beta}+r)^{\hat{\gamma}}}dr,
\end{equation}
while the relative measure is
\begin{equation}
\label{MR}
\mathcal{M}_R=\frac{\cal{M}_A}{\int_{1}^{31074}\frac{\hat{\alpha}}{(\hat{\beta}+r)^{\hat{\gamma}}}dr}.
\end{equation}

\section{Results and discussion}
\label{result}

\begin{figure}
\includegraphics[width=\textwidth]{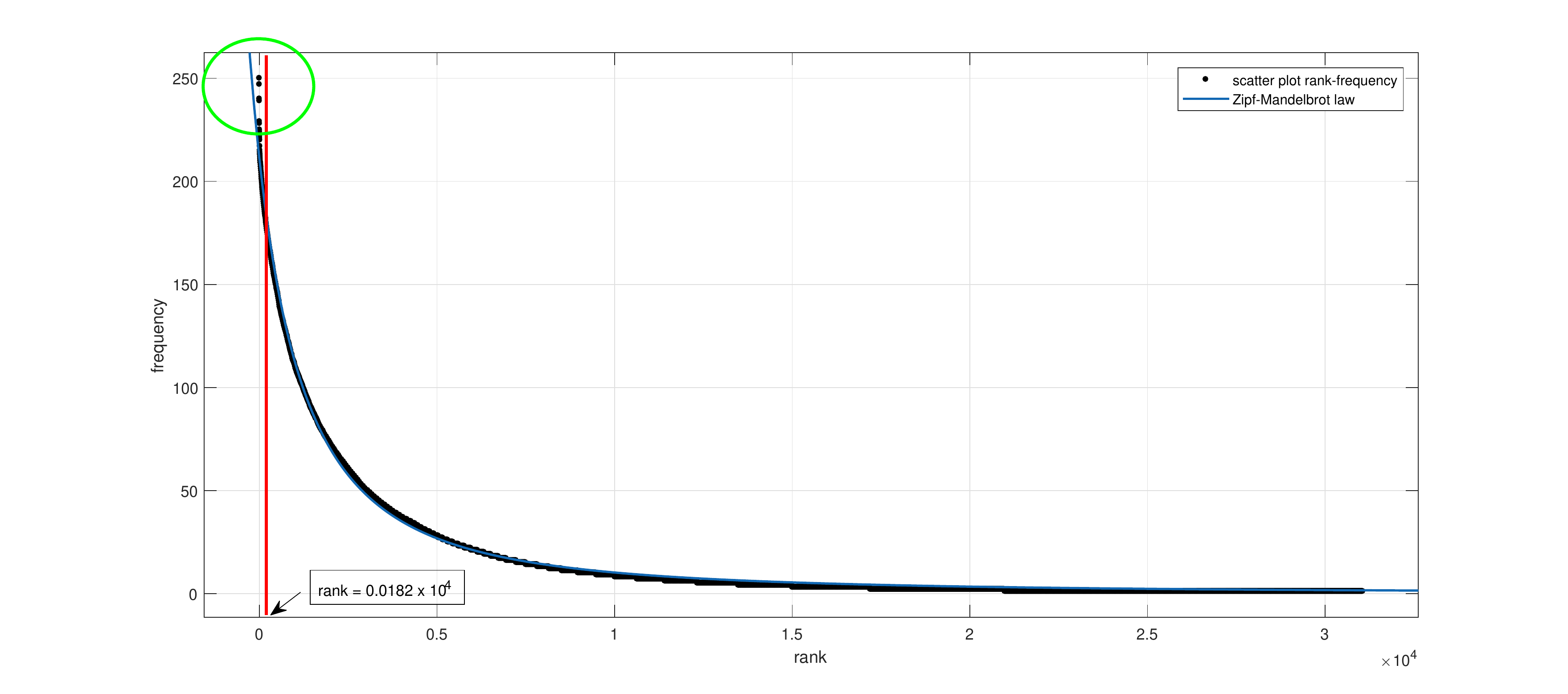}
\caption{Best-fit curve, according to equation (\ref{zipf}) and
calibrated parameters in Table \ref{Table1}, rows (a). The scatter plot of the
original sample is juxtaposed for a better comparison; the agreement
is very good; data and fits are hardly distinguishable from each
other. Notice the slight deviations at low ranks (green circle in
the Figure), suggesting  the presence of king and vice-roy effects
(see e.g. \cite{ACphysA}). The red vertical line points to
$\bar{H}=182$, which delimits the core of the
hapaxes.}\label{Fig1}
\end{figure}

The results of the best-fit exercise are reported in Table
\ref{Table1}, section (a), where one can find the calibrated parameters with the
confidence intervals at 95\%. The value of $R^2$ is 0.9971, which
suggests a quite perfect compliance of the considered ranked dataset
with the rank-size Zipf-Mandelbrot law. Figure \ref{Fig1} further
supports such a result by proposing a visual inspection of the fit.
A mere power (Zipf) law gives a much worse fit. 
\newline
In general, one can notice some kind of hidden pattern in the data when the fit is made by a power law. This is rather usual; at low rank, it is attributed to king and vice-roys effects; at large rank,  one attributes deviations to the fact that the frequency is low (=1 in our case) and because there is a plethora of  quantities (different words, in our case). This is usually much complicating numerical work,  for any type of theoretical and empirical fits  when there is such a high density of points but with low ``weight".  For information, there are more than 10000 (exactly 10088) ``absolute hapaxes"; for completeness let us mention that  there are  7574 ``relative hapaxes" appearing twice. Recall that there is a little bit more than  31000 data points. 

\begin{figure}
\includegraphics[width=\textwidth]{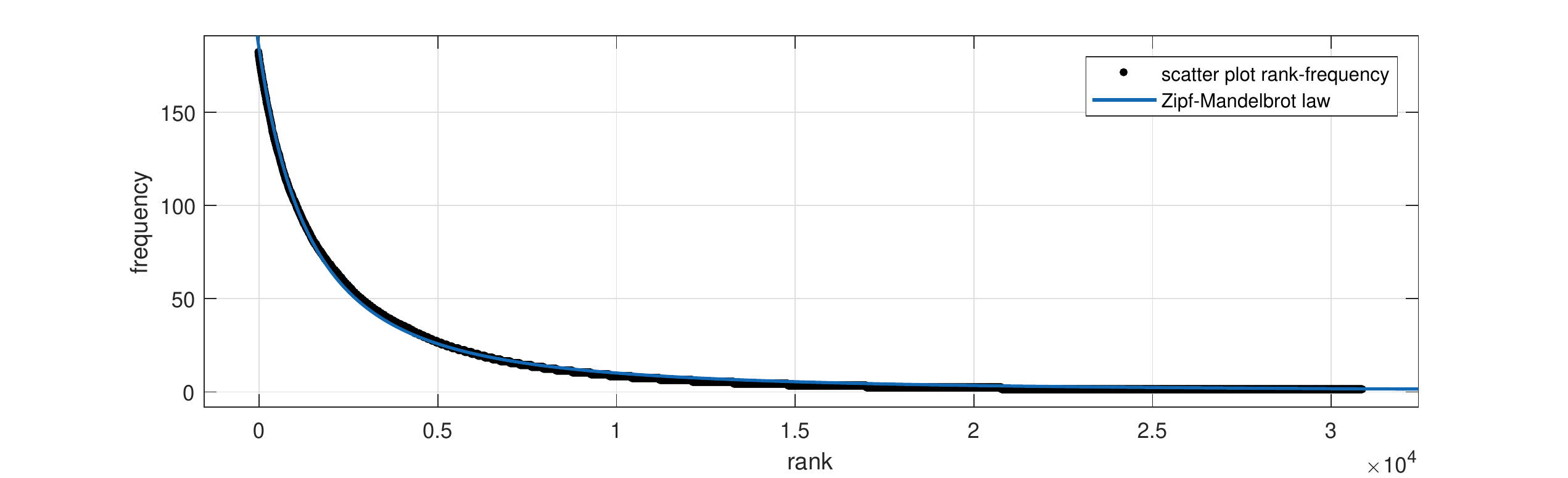}
\caption{Best-fit curve, according to Eq. (\ref{zipf}) and
calibrated parameters in Table \ref{Table1}, block (c), for the case of the
hapaxes excluding the core. The scatter plot and the fitted curve
are not distinguishable. The deviations at the low ranks shown in
Figure \ref{Fig1} do not appear, thus leading to the statement of
the presence of king and vice-roys effects for the elements of the
core in the respect of the overall sample.}\label{Fig3}
\end{figure}

That is why we have run the fit   reported in Figure \ref{Fig3}, with parameters reported in Table \ref{Table1}, rows (c). Such a figure provides also a very intuitive preliminary interpretation of the core of the hapaxes. Indeed, the removal of the core of the hapaxes from the overall ranked list lets the fit be quite perfect; this points to the claimed king and vice-roys effect mentioned above.

 \begin{table}
 \centering
 \begin{tabular}{c c c c c}
   \hline
	& & $\hat{\alpha}$ & $\hat{\beta}$ & $\hat{\gamma}$\\ \hline
   \multirow{2}{*}{(a)}   & Estimations  & $6.029 \times 10^8$ &        2540 & 1.896  \\\cline{2-5}
						   & Conf. Interv. 95\% & ($5.676\times 10^8$, $6.381\times 10^8$) &  (2525, 2554)
 & (1.890, 1.902) \\ \hline \hline
 \multirow{2}{*}{(b)} & Estimations & 287.7 & 5.903 & 0.084  \\\cline{2-5}
						& Conf. Interv. 95\% & (281.8, 293.6) &  (4.288,7.519) &(0.080,0.088) \\ \hline \hline
 \multirow{2}{*}{(c)} & Estimations & $4.359 \times 10^8$ & 2668 & 1.861  \\\cline{2-5}
					    & Conf. Interv. 95\% & ($4.083\times 10^8, 4.634\times 10^8$) &  (2652,2685) & (1.854,1.867) \\ \hline \hline
\end{tabular}
\caption{ (a) Best-fit parameters of the Zipf-Mandelbrot law in Eq. (\ref{zipf}) when all hapaxes are considered. (b) Best-fit parameters of the Zipf-Mandelbrot law in Eq. (\ref{zipf}) for the case of the hapaxes belonging to the core. (c) Best-fit parameters of the Zipf-Mandelbrot law in Eq. (\ref{zipf}) for the case of all the hapaxes without those belonging to the core. The ranges of the confidence intervals at 95\% for the three parameters are reported in parentheses for all the cases.}\label{Table1}
  \end{table}

We are now ready to treat the subsample given by the core of the hapaxes.
By looking at the data we have   $\bar{H}=182$, i.e. there
exist 182 words whose frequency is at least 182 and, simultaneously,
there are not 183 words with frequency at least 183. The most
frequent hapaxes are reported in Table \ref{most} for the reader
convenience. To save space, only one third of the core is shown,
i.e. the most frequent 61 hapaxes.

\begin{table}
  \centering
 \begin{tabular}{cc}
   \hline
   \textbf{Word(s)} &    \textbf{Frequency} \\ \hline
sense  & 250\\
given &  247 \\
bring, house & 240 \\
give & 239\\
hand, themselves & 229\\
within & 228 \\
others, therefore & 225\\
set & 224 \\
take & 222 \\
second  & 221 \\
find, full, making, since & 220\\
among & 217 \\
again, does, & 215 \\
itself, remain & 214 \\
being, brought, done, soon, whose &  213\\
part, protect & 212 \\
known, small & 211\\
able, beyond, carry, friends & 210 \\
call, day, far, fellow, means, opportunity, then, Washington, while
& 209 \\
course, order, single & 208 \\
essential, important, meet, reason & 207 \\
another, left, like, respect, seen & 206 \\
certain, few, necessary, possible, purpose & 205 \\
\hline
\end{tabular}
\caption{The most frequent 61 hapaxes, along with their frequencies. Other hapaxes have frequency larger than 200. Specifically, the positions/ranks between 62 and 80 have frequencies ranging in [204 - 200]. {Notice that different variations of the same word -- like ``give" and ``given" -- are considered as different words. This is in line with the employed methodological procedure (see Section \ref{data}).}}
\label{most}
  \end{table}

Thus, by applying formulas (\ref{MA}) and (\ref{MR}), and by using
the values listed in Table \ref{Table1}, section (a), a straightforward
computation gives that
\begin{equation}
\label{MAres}
\mathcal{M}_A=\frac{\hat{\alpha}}{-\hat{\gamma}+1}\left[(\hat{\beta}+\bar{H})^{-\hat{\gamma}+1}-(\hat{\beta}+1)^{-\hat{\gamma}+1}=
\right]=35783.9769 
\end{equation}
and
\begin{equation}
\label{MRres}
\mathcal{M}_R=\frac{(\hat{\beta}+\bar{H})^{-\hat{\gamma}+1}-(\hat{\beta}+1)^{-\hat{\gamma}+1}}
{(\hat{\beta}+31074)^{-\hat{\gamma}+1}-(\hat{\beta}+1)^{-\hat{\gamma}+1}}=0.0663 
\end{equation}

Notice that the hapaxes contained in the core represents a small
percentage -- about 0.58\% -- of the entire set of words said once.
However, in terms of frequencies, we have that the core is 6.63\% of
the overall set, as the relative measure assures. This means that a
very small set of words have been selected to be said only once in a
large number of speeches, with about eleven times the frequencies
over the hapaxes. One can conjecture that these are rare words but
purposefully intended.

\begin{figure}
\includegraphics[width=\textwidth]{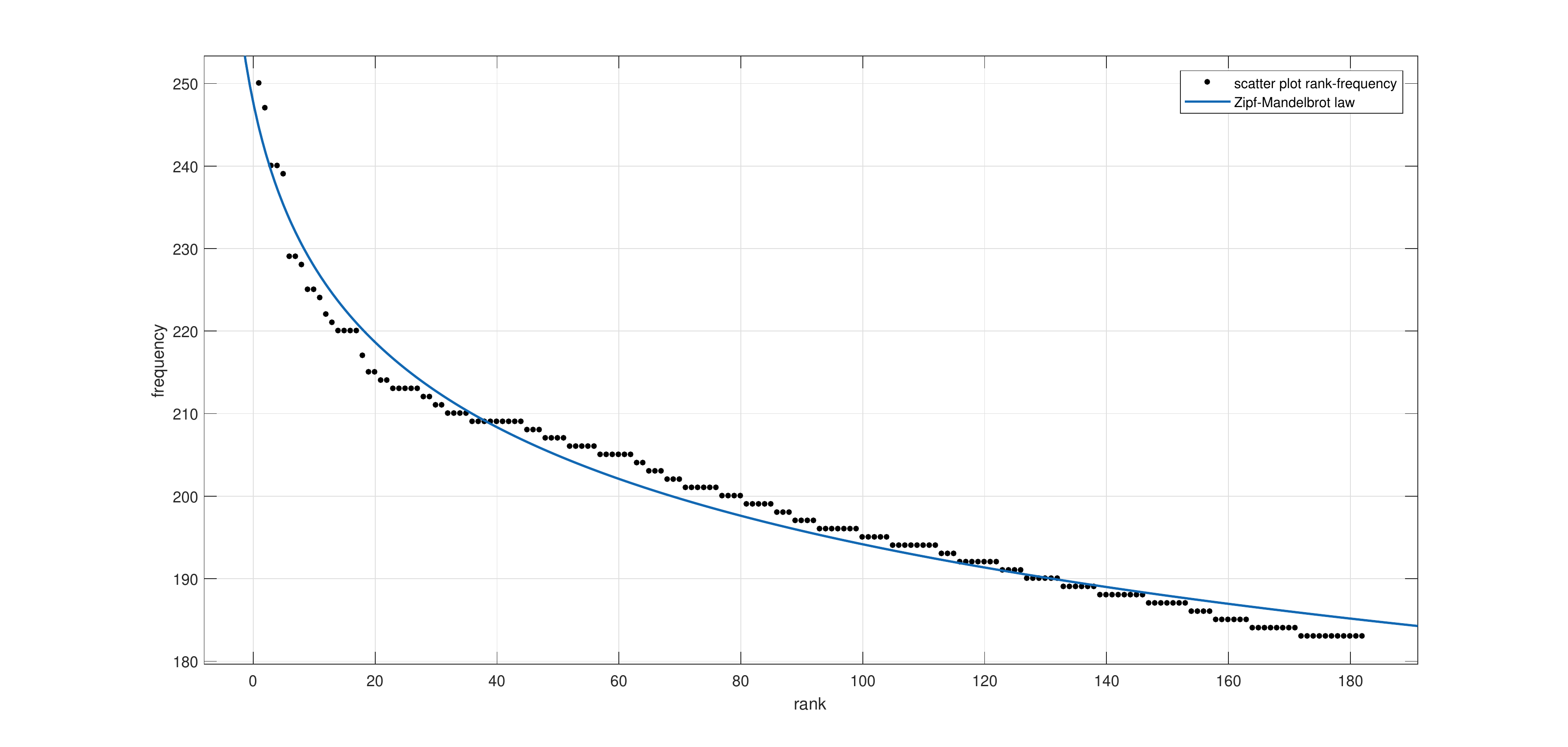}
\caption{Best-fit curve, according to Eq. (\ref{zipf}) and
calibrated parameters from Table \ref{Table1}, rows (b) for the case of the
hapaxes in the core. The scatter plot of the original sample of the
core is also shown for comparison purposes; the agreement is
visually good.}\label{Fig2}
\end{figure}

To have a view of the set of the core, we report in Table
\ref{stat} column (b), the main statistical indicators for the frequencies
of the set of such 182 hapaxes.
By exploring the core of the hapaxes itself, one can see that the
frequencies of the tokens therein contained represent a sample which
is well fitted by a Zipf-Mandelbrot law. Refer to Figure \ref{Fig2}
and Table \ref{Table1}, rows (b) for the details. The statistical goodness of
fit is rather satisfactory also in this case, with $R^2=0.978$. Also
the visual inspection suggests good compliance of the data with a
Zipf-Mandelbrot law, even if some evident deviations appear (see
Figure \ref{Fig2}). Such deviations are confirmed also by the wider
confidence intervals resulting in the case of the core with respect to those coming from
the overall sample, see Table \ref{Table1} rows (a) and (b).
Hence, similarly to what happens for the entire sample case (see Figure \ref{Fig1}), also in this case we have some points at the lowest ranks that are above the curve. Thus, the presence of the king and vice-roys effect is confirmed also when one takes such a subsample. The meaning of this outcome can be found in the presence of a few special words which have been pronounced only once in a large set of speeches. We can say that such words are \textit{necessary, but not to be stressed}. The five lowest ranked words in the core are \textit{sense, given, bring, house, give}. Their use is often ineluctable, but their abuse is not politically rewarding. Therefore, they appear often as hapaxes in a speech. For more details on the necessary, but not be stressed words, see the next subsection.

Figure \ref{Fig2} highlights also a queen and harem effect\footnote{Which is the respective of king and vice-roys but manifested at the highest ranks} at the highest ranks. Indeed, the hapaxes of the core which appear more rarely in the US Presidents' speeches are below the best fit curve. This outcome confirms the linguistic evidence that the less frequent 
hapaxes of the core are relatively less scattered than the more frequent ones. Furthermore, the elements of the core with highest ranks form a subset of \textit{identically pronounceable words resulting to be hapaxes}; specifically, such words form a collection of tokens which have been used in the same number of occasions by the US Presidents. The queen and harem effect appears because the set of such identically pronounceable words used as hapaxes has a particularly large cardinality. In noting this, we provide further confirmations that the US Presidents use words as devices for communication strategies, having a clear idea on the proper terminologies -- always the same -- to be used only once in specific situations. This goes in the direction of stating a substantial tendency of the politicians to imitate the past when delivering a talk, as already discussed in the paper.

The presence of king and vice-roys plus queen and harem effects distort the best fit curve, which then is constantly below a group of middle-rank hapaxes of the core (see again Figure \ref{Fig2}). Thus, such a special set of elements of the core -- denoted hereafter as \textit{regime of the highly frequent middle-rank hapaxes} -- has a behavior which is analogous to that of the identically pronounceable words used as hapaxes. In fact, the existence of the regime of the highly frequent middle-rank hapaxes is due to the properties of such words to be pronounced as hapaxes about the same number of times. Specifically, hapaxes in this set have the same size or there are no noticeable changes in size when the one moves from a rank to the subsequent one. Therefore, one can offer an interpretation of the regime of the highly frequent middle-rank hapaxes which is similar to that of the set of identically pronounceable words employed as hapaxes.

Notice that the cardinality of the sample set is able to affect the goodness of
fit; in particular, in some circumstances, one can claim that larger
cardinality leads to less scattered data.

%


The hapaxes in the core produce a king (the word \emph{sense}, with
frequency 250) and 181 vice-roys effect. Indeed, once the core is
removed from the sample, one obtains a perfect fit through a
calibrated Zipf-Mandelbrot law, because of the removal of the deviations
at the low ranks (compare Figures \ref{Fig1} and \ref{Fig3}).
Similar  deviations at low rank can be found in studies on other
types of data, e.g. city size, or co-author distributions (see e.g.
\cite{ls, ausloos2013, ACphysA}). In the case of core removal, the
goodness of fit remains quite perfect, with $R^2=0.9965$.
The best fit parameters can be found in Table \ref{Table1}, rows (c), along
with the related confidence intervals.


%


%

\subsection{Implications of the analysis: a discussion of some salient hapaxes} \label{hapaxdiscussion}
\label{hapaxexplanation}
As a preliminary premise, we argue that the most evident implication of our proposal lies in the identification of relevant hapaxes (those in the core) whose meaningfulness has to be discussed also in the light of their role in the context.
\newline
Indeed, it is important to claim that a full and complete description of the connection between the single words in the core and the US Presidents communication strategy can be properly discussed only by involving a semantic and local analysis of the tokens -- along with their contextualization. In saying so, we claim that this paper can be viewed as the presentation of a promising methodology and the starting point of future studies in the linguistic arena.
\newline
We here provide some suggestions leading to a more detailed interpretation of the results of the analysis. In particular, a discussion of some of the most frequent hapaxes resulting from the analysis is reported here to show the potentiality of the presented approach. Such a discussion proceeds in two steps: first, we give some general insights on the content of the core of the hapaxes, with a special attention to some classes of such tokens; second, we provide a detailed analysis of some specific highly recurrent hapaxes, along with a discussion on their contextualization in noticeable speeches. We have also implemented a specific poly-grams\footnote{A poly-gram is a contiguous sequence of $n$ words from a speech transcript.} analysis to explore the context in which the hapaxes occur. The phrases reporting the terms of interest are reported and commented in the next rows.
\newline
Still in the line of highlighting the implications and the applications of our approach, we also report a discussion of some hapaxes appearing strictly once in the entire corpus.
\newline
In the core, one can find the following terms: \textit{opportunity, purpose, confidence, progress}. Such words suggest a positive reading of the subject of the speeches. Since they are hapaxes, one can reasonably argue that they are associated to a substantially negative situation -- hence, they are pronounced only once -- which the Presidents try to revert into a positive perspective. This communication strategy has appeared several times in the history of the US Presidents speeches. For example,  \textit{``And our neglected inner cities will see a rebirth of hope, safety, and \underline{opportunity.}" is the phrase pronounced by Donald Trump in the Address to Joint Session of Congress on February 2017.
\newline
Another class of nouns collects \textit{duty, efforts, honor, justice, responsibility, effort, others, friends,
respect, order, safety, faith, hope, hands}. Here the Presidents appeal to the sentiments of the auditors and call for empathy and political support. Also in this case, intuition suggests that the content of the messages containing such hapaxes was related to troubling situations. In this worrying context, the brief reference to the worthiness of the citizens can be viewed as an effective communication device against pessimism. For example,  from the aforementioned talk stated by Trump on February 2017:  \textit{``Also with us are Susan Oliver and Jessica Davis.  Their husbands, Deputy Sheriff Danny Oliver and Detective Michael Davis, were slain in the line of \underline{duty} in California.  They were pillars of their community.  These brave men were viciously gunned down by an illegal immigrant with a criminal record and two prior deportations.  Should have never been in our country."} or the State of the Union Address from Bill Clinton on February 1997, where the President said: \textit{``We have much to be thankful for. With four years of growth, we have won back the basic strength of our economy. With crime and welfare rolls declining, we are winning back our optimism, the enduring \underline{faith} that we can master any difficulty. With the Cold War receding and global commerce at record levels, we are helping to win an unrivaled peace and prosperity all across the world."}
\newline
In a similar context, the introduction in the speeches of \textit{protect, respect, support} as hapaxes points to the efforts spent by the Presidents in defending the auditors, even in presence of adversities. For example, President Obama in the Speech on the Strategy in Afghanistan and Pakistan in December 2009 said: \textit{`` For the first time in its history, the North Atlantic Treaty Organization invoked Article 5 -- the commitment that says an attack on one member nation is an attack on all.  And the United Nations Security Council endorsed the use of all necessary steps to respond to the 9/11 attacks.  America, our allies and the world were acting as one to destroy al Qaeda’s terrorist network and to \underline{protect} our common security."}
\newline
We now point our attention to a set of adjectives which belong to the core of the hapaxes: \textit{essential, important, necessary}. Such tokens can be associated to the political will of the speakers when referring to the ineluctability of an intervention which presents potentially negative outcomes. For example, President Obama talking to the UK Parliament, just after mentioning the joint intervention in Libya in 2011, said: \textit{``Our action - our leadership - is \underline{essential} to the cause of human dignity. And so we must act - and lead - with confidence in our ideals, and an abiding faith in the character of our people, who sent us here today."}
\newline
The classes of hapaxes of the core introduced above allow to have a general view of the way in which US Presidents and their collaborators have created the official speeches. Some communication strategies appear with a certain degree of recurrence, and this goes in the direction of explaining the presence of an imitative behavior among the Presidents. 
\newline
We are now ready to clarify some more specific regularities in the Presidential talks by analyzing the patterns of some peculiar hapaxes.}
\newline
Let us start checking which Presidents have used the most common hapax \textit{sense} (see Table \ref{senseperc}, left box) with a specific focus to the reference context. Four Presidents (John
Adams, Martin Van Buren, Zachary Taylor and Rutherford B. Hayes)
noticeably used ``sense" as hapax. They employed it in circumstances
of celebrations or very formal contest like Annual Messages,
Proclamations or Veto announcements. The refine rhetoric used at the
time of the aforementioned Presidents (1797 - 1883) jointly with
particularly important aforesaid appointments of Presidents'
political agenda, create the ground for the use of the hapax ``sense"
to qualify feelings, with references to a common perception of
something. For example, Chester A. Arthur, in ``Veto of River and
Harbors Act " of August, 1882 said: \textit{`It is not necessary
that I say that when my signature would make the bill appropriating
for these and other valuable national objects a law it is with great
reluctance and only under a \underline{sense} of duty that I
withhold it.'}. Chester A. Arthur has used ``sense of duty" in the
circumstances of a Veto Message, when he was trying to convince his
audience about his arguments\footnote{\url{www.u-s-history.com/pages/h735.html}}, but
the Congress has overridden the veto and the legislation was
approved.
\newline
John Adams, Martin Van Buren, Zachary Taylor and Rutherford B. Hayes
have evoked sentiments like \textit{`\underline{sense} of national
honor, dignity, and independence '} (John Adams; December 8, 1798:
Second Annual Message), \textit{`the good \underline{sense} and
patriotism'} (Martin Van Buren; December 3, 1838: Second Annual
Message to Congress) or \textit{`\underline{sense} of the duty'}
(Rutherford B. Hayes; December 1, 1879: Third Annual Message)
referring to specific common feelings. From 1900 on, the bigram
``common sense" has been employed 29 times out of 250 times that
``sense" appears as hapax. The Presidents who have evoked most
frequently ``common sense" are Ronald Reagan and Franklin D.
Roosevelt. The former points to the common sense in formal contexts
like State of the Union Addresses of January, 1988 and February,
1986, while the latter appeals to common sense during his famous
fireside chats (Fireside Chat 5: On Addressing the Critics, Fireside
Chat 10: On New Legislation, Fireside Chat 15: On National Defense).
Bear in mind that the fireside chats have been instituted by the
President to have a colloquial level of communication with citizens.
Here we report some words pronounced in the conclusive part of the
``Fireside Chat 10: On New Legislation" of October, 1937:
\textit{`The common \underline{sense}, the intelligence of the
people of America agree with my statement that ``America hates war.
America hopes for peace. Therefore, America actively engages in the
search for peace." '}. This part of the discourse was about
peace. However, the President mentions ``common sense" once, to
deliver the (hidden) message that US citizens should exhibit
responsibility and unity in the terrible case of non avoidable war
(that was the case, indeed, as the President probably guessed.
Remember that Hitler had re-militarized the Rhineland in March 1936,
just before the speech). Thus, hapax could be a part of a bigger
rhetoric framework adopted by Franklin D. Roosevelt to prepare the
public opinion for future military actions without explicitly
mentioning them. Substantially, Presidents are aware that the bigram ``common sense"
might be efficiently used to introduce wisdom and sustainability
criteria in military and engineering contexts.
\newline
Concluding, the presence of ``sense" as the \textit{king} of the
hapax list can be justified by assuming that the
speakers$\backslash$writers have paid attention in referring to any
type of common sense when dealing with public communications, even
if without stressing it too much. Indeed, when a public speaker is
invoking a shared sense of something, he has to bear in mind that,
what is considered common sense for him or for a certain
sub-community of auditors, is not common sense for others, so using
certain hapax locutions just once, he can send a specific message to
a particular group (characterized by the appropriate sensibility
needed to receipt that particular message) or he is trying to keep
people feeling the sense that he is appealing to, in order to
inspire a specific collective behaviour, e.g. see \cite{Ivie1984}.
\newline
A second interesting hapax is \textit{bring}. From Table
\ref{senseperc} right block, emerges that Presidents Richard Nixon, Harry S.
Truman, Chester A. Arthur and John Tyler are the Presidents most
often utilizing the hapax ``bring". The aforementioned Presidents did
not employ ``bring" in colloquial situation, at least in the speeches
stored in this dataset. The hapax occurs in formal and relevant
political events to manifest the immediate need of an action, or to
describe the effects of a full willingness to act in a certain
direction. In particular, the hapax ``bring" has a relevant use in
relation to tensions, conflict or war. For example, in ``Address to
the Nation on Presidential Tape Recordings" of April 1974, Richard
Nixon said \textit{`These conversations are unusual in their subject
matter, but the same kind of uninhibited discussion—and it is
that—the same brutal candor is necessary in discussing how to
\underline{bring} warring factions to the peace table or how to move
necessary legislation through the Congress.'} referring to the
political struggle started with the Watergate case. Another salient
example is the speech of September 1945, ``Announcing the Surrender
of Japan" stated by Harry S. Truman, where he said \textit{`No
victory can bring back the faces they longed to see.'} referring to
the deaths of that war; or the speech of September 1948,
``Whistlestop Tour in Trenton, Missouri" during which Harry S. Truman
said: \textit{`There is one thing I want to \underline{bring} home
to you.'}. This comment can be extended to all the other Presidents
that have used ``bring" as hapax: indeed the word ``war" occurs 37
times within the phrases in which ``bring" appears.
\newline
The behaviour of ``bring" as hapax along the Presidents is different
from the previous one, (compare columns in Table \ref{senseperc}). The ``bring" occurrence is more homogeneously
distributed. Consequently, it is more difficult to grasp dramatic
changes along the presidencies, but we can note that the hapax
``bring" has a longer life, being used from November 6, 1792, by
George Washington in his ``Fourth Annual Message to Congress" to very
recent speeches.
\newline
The hapax legomenon \textit{house} has a peculiar regularity in
occurring, because it is commonly used in the introductory
statements of many ceremonial moments of the US political agenda.
For example, the introductory forms `To The Senate and House of
Representatives' and `Fellow Citizens of the Senate and of the House
of Representatives' are the source of the usage of ``house" as hapax
for 53 and 26 times respectively (a poly-grams comparison has been run to figure it out). These forms were commonly used
between January, 1790 (``First Annual Message to Congress" of George
Washington) and December, 1932 ( ``Fourth State of the Union Address"
of Herbert Hoover). The utilization of such locutions has originated
from the fact that during the aforementioned period, most of the
messages were not personally stated by the Presidents, but they were spread
in written form (Thomas
Jefferson, Woodrow Wilson and Franklin D. Roosevelt are the unique
that have not respected this common practice). According to \cite{Kolakowski2006}, Franklin D. Roosevelt established the personal appearance as a
permanent tradition with his 1934 State of the Union Message. After
that talk, the so called ``President’s Annual Message to Congress"
starts to be known as ``State of the Union Address". Those points are
confirmed by our findings. The two introductory locutions listed
above were employed mainly for Annual Messages and State of the
Union Addresses until 1932, but later, under the Franklin D.
Roosevelt Presidency, they stopped to occur. After that, the use of
``house" is mostly associated to the bigram ``white house" which
justifies the use of the hapax for 64 times out to 240. In the light
of this evidence, we point out that the employment of ``white house"
as hapax has been intensified after Harry S. Truman, whose
Presidency is the one during which the White House has been restored\footnote{\url{http://www.whitehousemuseum.org/special/renovation-1948.htm}}.
Furthermore, thank to a visual inspection of the context in which
``white house" has been used, it results that the Presidents refer to
it for indicating more than a mere physical place (it is a case of
\textit{metonymy} utilization). They address to the White House for
pointing to the residence of the US most representative political
institution. For example, on March 21, 2013, during the ``Address to
the People of Israel", Barack Obama said: \textit{`Just a few days
from now, Jews here in Israel and around the world will sit with
family and friends at the Seder table, and celebrate with songs,
wine and symbolic foods. After enjoying Seders with family and
friends in Chicago and on the campaign trail, I’m proud that I've
now brought this tradition into the White \underline{House}. I did
so because I wanted my daughters to experience the Haggadah, and the
story at the center of Passover that makes this time of year so
powerful'}. Such a speech highlights the importance of having
certain events into the White House, and the hapax implicitly states
the relevance for all the Americans.
\begin{table}[ht]
\centering
\begin{adjustbox}{totalheight=\textheight-2\baselineskip}
\begin{tabular}{lrr||lrr}
\hline
\addlinespace[0.1cm]
Presidents & \textbf{sense [\%]} & tot. speech & Presidents & \textbf{bring [\%]} & tot. speech\\
\hline
\addlinespace[0.1cm]
John Adams & 66.67 & 9 &Richard Nixon & 39.13 & 23 \\
Martin Van Buren & 60.00 & 10 &John Tyler & 38.89 & 18 \\
Zachary Taylor & 50.00 & 4 &Harry S. Truman & 36.84 & 19 \\
Chester A. Arthur & 45.45 & 11 &Chester A. Arthur & 36.36 & 11 \\
Rutherford B. Hayes & 43.75 & 16 &Gerald Ford & 35.71 & 14 \\
William Taft & 41.67 & 12 &Andrew Johnson & 35.48 & 31 \\
Ronald Reagan & 36.84 & 57 &George H. W. Bush & 35.00 & 20 \\
Barack Obama & 36.00 & 50 &Bill Clinton & 34.21 & 38 \\
William McKinley & 35.71 & 14 &Abraham Lincoln & 33.33 & 15 \\
John F. Kennedy & 34.15 & 41 &Calvin Coolidge & 33.33 & 12 \\
Dwight D. Eisenhower & 33.33 & 6 &Dwight D. Eisenhower & 33.33 & 6 \\
George Washington & 33.33 & 21 &Jimmy Carter & 33.33 & 18 \\
Warren G. Harding & 33.33 & 18 &John Adams & 33.33 & 9 \\
Franklin D. Roosevelt & 32.65 & 49 &Ulysses S. Grant & 31.25 & 32 \\
James Monroe & 30.00 & 10 &Barack Obama & 30.00 & 50 \\
Bill Clinton & 28.95 & 38 &James Monroe & 30.00 & 10 \\
Lyndon B. Johnson & 28.79 & 66 &Martin Van Buren & 30.00 & 10 \\
Gerald Ford & 28.57 & 14 &Franklin D. Roosevelt & 28.57 & 49 \\
James Buchanan & 28.57 & 14 &Millard Fillmore & 28.57 & 7 \\
George W. Bush & 28.21 & 39 &George W. Bush & 28.21 & 39 \\
James K. Polk & 28.00 & 25 &James Madison & 27.27 & 22 \\
Woodrow Wilson & 27.27 & 33 &Ronald Reagan & 26.32 & 57 \\
Andrew Jackson & 26.92 & 26 &Lyndon B. Johnson & 25.76 & 66 \\
Herbert Hoover & 23.33 & 30 &Zachary Taylor & 25.00 & 4 \\
Richard Nixon & 21.74 & 23 &Woodrow Wilson & 24.24 & 33 \\
Benjamin Harrison & 21.05 & 19 &George Washington & 23.81 & 21 \\
Franklin Pierce & 20.00 & 15 &Thomas Jefferson & 20.83 & 24 \\
George H. W. Bush & 20.00 & 20 &Franklin Pierce & 20.00 & 15 \\
Grover Cleveland & 17.24 & 29 &John F. Kennedy & 19.51 & 41 \\
Calvin Coolidge & 16.67 & 12 &Andrew Jackson & 19.23 & 26 \\
John Tyler & 16.67 & 18 &Rutherford B. Hayes & 18.75 & 16 \\
Andrew Johnson & 16.13 & 31 &Grover Cleveland & 17.24 & 29 \\
Harry S. Truman & 15.79 & 19 &Herbert Hoover & 16.67 & 30 \\
Millard Fillmore & 14.29 & 7 &James K. Polk & 16.00 & 25 \\
James Madison & 13.64 & 22 &James Buchanan & 14.29 & 14 \\
Abraham Lincoln & 13.33 & 15 &John Quincy Adams & 12.50 & 8 \\
John Quincy Adams & 12.50 & 8 &Benjamin Harrison & 10.53 & 19 \\
Ulysses S. Grant & 12.50 & 32 &William Taft & 8.33 & 12 \\
Theodore Roosevelt & 9.09 & 22 &William McKinley & 7.14 & 14 \\
Thomas Jefferson & 4.17 & 24 &Warren G. Harding & 5.56 & 18 \\
\hline
\end{tabular}
\end{adjustbox}

\caption{The percentage of the speeches per President that contain the word ``sense" and ``bring" (respectively left and right boxes) as hapaxes. The sub-tables are ranked according to ``sense [\%]" and ``bring [\%]" respectively. } \label{senseperc}
\end{table}

Additionally, to do the best for providing insights on the worthiness of the analysis of the words pronounced only one time, we now discuss some hapaxes appeared just once in the whole corpus. Their meaning is relevant for the political context even if they are not belonging to the core of hapaxes.
\newline
The names \textit{Dostoyevsky, Kandinsky, Scriabin, Uzbek, Alisher Navoi, Boris Pasternak and Zhivago} have been stated by the President Reagan on the May 31, 1985 during his `Address at Moscow State University'. They are good examples of meaningful hapaxes. Indeed, they occurred all together when the President Reagan was trying to emphasize his appreciation and familiarity with the Russians' culture. He has used them for giving strength to a speech mostly dedicated to ``freedom'' and ``truth'' just few month before the fall of the Berlin wall.
However, it was a local phenomenon confined in a specific situation. Differently, the cases of ``sense'', ``bring'' and ``house'' represent something of a more general contextualization; in fact, they belong to the core of hapaxes. This is expected because the threshold calibrated on the H-index helps in identifying a list of hapaxes whose messages are constantly present across the talks. Summarizing, our proposal catches unit of information (words) due to their degree of systematic singular usages. Consequently, it is easier to highlight the delivered latent messages; this leads to the reading of the US political communication history under a different perspective.

\subsection{Extension to poly-grams: some remarks}
\label{poly-grams}
A relevant theme to be discussed is the extension of the analysis to poly-grams said only once in each speech. Indeed, bi-, tri- and, in general, poly-grams might turn out to be useful for grasping further information on the political speeches. However, there is no room here to face this aspect in an exhaustive way. We here elaborate on this relevant problem.
\newline
First of all, we have arguments for thinking that the exploration of poly-grams is quite much demanding from an analytical point of view. In fact, the one word case can be treated by considering a token as a unique set of consecutive letters, and two tokens as divided by blank spaces or punctuation. Bi-grams are two consecutive tokens considered together, and the distinction rule applied for one word is not longer applicable in this case. Therefore, the shortest meaningfulness unit of analysis has to be the one-word token, hence leading to the evidence that the main driver of the sense of a poly-gram would remain the same we are considering here for one word.
Moreover, the number of bi-grams said only once should evidently be much higher than the number of hapaxes, and the number of tri-grams said once should be greater than the case with two words. Such a tendency goes on till the point in which the number of words of the poly-gram is small enough. Indeed, in the extreme case of one hundred words poly-gram in a speech with one hundred words, then the entire speech can be seen as the only poly-gram said only once. However, when aggregating over all the speeches, the growing number of cases to be treated for bi- or tri-grams may lead to an extremely complex computational procedure, specially for large corpus, and this can turn out to be a severe drawback of the methodological analysis of the problem.
\newline
By a completely different perspective, a relevant issue to be carefully considered concerns the meaning of the poly-grams. One-grams are words with certain meaning; they can be considered, as argued here, as if bringing purposeful ideas. The meaningfulness of the poly-grams is somewhat questionable. One can have sequences of words which are forced by grammatical and syntactical constraints, or poly-grams whose logical sense cannot be appreciated when taken out from the context. As an example, if we consider the bi-gram "America is", the selected word is "America", while the term "is" is a trivial grammatical constraint. The President might also select "Our country is". The real difference lies in the choice between "America" and "country". In the former case, attention is paid explicitly to America, while in the latter one the President points to a more general term like "country". It is also worth to observe that if "America" is a hapax, then "America is" is a bi-gram said only once. The converse is not true in general, and even if "America is" is a bi-gram said only once, the word "America" is not necessarily a hapax. So, even if different information are captured, there will be the need of a reading phase to detect those poly-grams really referring to a recurring marginal topic in the US President speeches. 
\newline
As a further consideration, one can easily see that there is a critical set of words which are associated to poly-grams said only once in a specific speech. The number of words in such a critical set does not increase as the length of the speech decreases. As an intuitive example, if we take a sentence with ten words, most likely all the bi-grams are never repeated more than one time. In this respect, we observe that the length -- in term of words -- of a generic US Presidents' speech is rather small. Thus, one may likely have that all the poly-grams composed by a small number of consecutive words are said only once. Such remarks suggest that the outcomes of the study of the poly-grams can present some sources of biases. We will go back to this point in the next Section, when discussing potentially interesting research topics.

\subsection{Conjectures and themes for a discussion from the linguistic perspective}
\label{conj}
Under the point of view of the linguistic field, we acknowledge the evident discrepancies between political and other types of texts -- like scientific ones. In so doing, we aim at providing some conjectures coming from the study. 
\newline
Indeed, scientific speeches can be of two very different categories. By one side, scientists deliver speeches to illustrate the content of their studies to the scientific community. Appropriate occasions for this type of talks are conferences and seminars at Academic Institutions; by the other side, a scientist can pursue the scope of divulging her/his scientific achievements to people not involved in the field. In this case, the talk can be delivered during a conference press or in non scientific meetings open to the public.
In this latter category, one can find also the situation in which scientists inform politicians and policymakers on the results of research activities having an impact on the society -- hence, calling for regulation and political actions. 
\newline
The scientific speeches addressed to a scientific community have a structure which is structurally biased by their contents. Indeed, researchers are constrained to employ some specific words when referring to peculiar concepts, to maintain the level of the text as rigorous as possible. Basically, synonyms do not  appear often in pure scientific talks. As an example, we can mention the derivative of a (regular enough) function. It is possible to refer to the derivative by means of a formula, but scientists do not have a different term to denote such a mathematical concept. Therefore, when the content of the study includes the derivative of a function, the speaker is forced to employ always the word \textit{derivative} when referring to it.
\newline
A completely different perspective is associated to scientific talks aiming at disseminating a research achievement. In this case, the authors of a scientific study are obliged to avoid technical terms; hence, they can take the advantage of expressing scientific concepts by means of words of common use. In so doing, researchers are able to properly build their speeches in the light of a pre-selected communication strategy. 
\newline
The arguments expressed above suggest that political speeches are quite similar to scientific speeches, when the latter ones are of divulging type. If the scientific speech has a purely academic and technical nature, then the selection of a peculiar linguistic structure or the decision of employing some particular hapaxes is quite out of the range of the possibilities of the speaker.
\newline
In this respect, it is important to mention \cite{parsons}, where the Author elaborates on the communications between politicians and scientists. Specifically, the former category is claimed to serve as a target audience for divulging the outcomes of scientific research. Indeed, as pointed out above, it is crucial that politicians understand the most relevant scientific contexts. The quoted paper clearly illustrates that scientific divulgation and political messages should follow similar communication patterns in order to obtain such a desired target. 
\newline
Thus, one car argue that researchers have to imitate politicians in terms of communications, if they aim at letting their studies be understandable by the policymakers. Of course, this represents a difficult task for scientists.
\newline
To conclude, a detailed study is expected to offer a high level of discrepancy when comparing the linguistic structures of scientific speeches and political ones. Such deviation would tend to reduce when researchers pursue the target of stimulating the action of the politicians. In term of hapaxes, the scientific speeches aimed at divulging science should present a core of hapaxes made of non-veiled topics but rather the core will contain technical words demanded by the context. Therefore, to produce a meaningful analysis with scientific talks, one should get a homogeneous corpus, for example the Nobel Prize speeches. 
\newline
Under a more general perspective -- but still in the linguistic arena -- one can advance some further conjectures representing additional noticeable topics for future research. Such conjectures provides also a motivation for the present study.
\newline
First, we conjecture that a word said only once is not always delivered to
all, but it might be pronounced to reach a qualified group of
auditors who are able to grasp the message. In this respect,
evidence suggests that a word pronounced several times rings the
bells of the widest part of the audience, and sometimes the
comprehension of the mass is viewed as a negative outcome of the
speech; second, we conjecture that hapaxes might reflect concepts whose explanation is
absolutely necessary, but associated to situations that the
President recalls reluctantly to the audience. In this sense, in
order not to be attracting too much the attention of the auditors to a
specific argument -- whose mention is, however, necessary -- the
Presidents might wish to pronounce some words only once;
third, we conjecture that hapaxes might be related to situations and concepts
that should not be included in a speech and, despite this fact, are
intentionally mentioned by the President. Intuitive examples of such
cases are the ones related to diplomatic accidents. In such
situations, the annoyance of a President cannot be strongly
stressed, in the light of maintaining excellent relationships with
institutions and commercial partners. However, the President cannot
be silent in the regard of an insult or of an improper behaviour, in
order to not lose credibility. The hapax is the right mean for these
types of situations.
\newline
Furthermore, the analysis of the frequencies of the hapaxes over a large set of
speeches -- which is exactly our methodological ground here -- provides additional insights. Indeed, all the conjectures listed
above should be interpreted in the more complex environment of
several Presidents and occasions. This remark leads to some (further) plausible research questions.
In particular, it is worth noting that largely pronounced hapaxes raise a question about the presence of recurrent themes, treated by the President(s), also in the light of the conjectures listed above. Furthermore, the recurrence of a hapax raises a question about the
presence of an imitative behaviour of the Presidents to their
predecessors. Substantially, there is evidence of the will to learn
from the past in the communication strategies.
It is also important to point out that the employment of specific hapaxes by several
Presidents raises a question about the existence of a sort of linguistic code.
Finally, since hapaxes are rather rare words, one can conjecture that they are taken from a dictionary black box, only containing the never used words.

\section{Conclusive remarks}
\label{conclusioni}
This paper faces the challenging theme of exploring part of the
content of the official political speeches with an innovative method. The paradigmatic case of US Presidents' talks serves as a guide.
We start from the premise that official speeches are carefully written because the messages carried out are highly influential.
Each talks contains information to be delivered to some audience and aimed at the entire society. Thus, words are tactically selected with care, depending on the situation.
\newline
We are interested in the hapaxes of each speech, which are relevant units of information in the Presidents communication strategy.
In fact, one can observe some recurrent hapaxes in the corpus. They have consciously  pronounced / selected only once in several occasions and by several Presidents.
The relative rarity of these is thought to be intentional, sometimes appearing as new (or astute) words,
implying the President modernity, elitism, and wide knowledge. Anyway, we consider the approach here designed as promising for studying even much larger corpora that spreads across many years.
\newline
If appropriately merged and ranked, hapaxes show regular paths and
can be successfully fitted by the Zipf-Mandelbrot law. Moreover, there is a privileged set, the core hapaxes, here defined through the introduction of a Hirsch-based threshold.
\newline
We have shown that a small number of words have been pronounced
once per speech several times in official communications. This confirms and lets us
understand the presence of common messages and arguments in the
historical paramount view of the US Presidents' interventions. This list represents the core of the hapaxes. Such a core can be interpreted as those words which strike a point, even though
they are rarely used within each text of the corpus.
\newline
We have also shown that the core has a structure similar to the one
of the overall sample, with a compliance with a rank-size law of
Zipf-Mandelbrot type. Moreover, the core is also responsible for deviations of the overall
set of hapaxes from the best Zipf-Mandelbrot curve. In this, king
and vice-roys effects are detected.
\newline
The analysis of some relevant hapaxes is presented in Section \ref{hapaxdiscussion}, to illustrate the implications of the new approach. There we have looked for the contexts in which certain hapaxes occurred, justifying their presence. This is fundamental to understand the potential of our approach.
\newline
From Section \ref{result} the ability of our method in responding to the research questions clearly emerges. Indeed, the hapax legomena occurring in a collection of texts can be huge and their meanings are remarkable, as already stressed above. One should go trough each of them manually and understand the context in order to decide which one are connected to a global phenomenon and which are linked to a local one. 
\newline
It is important to notice that this study cannot offer a systematic analysis of all the hapaxes which occurred a few times (and, in particular, just once) in the whole corpus of the US Presidents' speeches, because their contextual meaningfulness in the respective speeches may lead to a very historic / social specific context, a spurious results or to a somewhat questionable informative content. 
In this respect, it is worth noting that in defining the core of hapaxes and a procedure to determine it, we filter out tokens until we get a list of words which have been regularly used by the Presidents;  therefore, the words in the core of the corpus have passed a selection process through years of usages and political phases, sometimes ending up to be pronounced just once per speech. On the other hand, certain hapaxes occurring just one time in the corpus may be part of a very local event. To be able to capture it, a reading inspection performed by an expert like a philologist rather than an automated procedure is required; this is well-beyond the target of the present paper. 
Differently, in the case of those words belonging to the core, the likelihood to be relevant is much higher.
\newline
Finally, the presented analysis is fully reproducible in different
contexts and is particularly useful when a researcher is facing a
large collection of documents coming from a uniform environment and
difference sources. Indeed, if s/he wants to easily access the
presence of marginal topics or remarkable semantic outliers, the
analysis of the core of the hapaxes can be of special usefulness
(see \cite{Greenspahn2016} as example of an extensive analysis of
hapaxes and the justification of such an investigation).
Furthermore, our procedure supports the assessment of a
cross-document view, which is helpful for the identification of
latent marginal keywords that may point to a \textit{fil rouge}
between texts.
\subsection{Future research}
We observe that the present study represents a further step towards the methods to investigate corpora; it is a step ahead in the comprehension of the political speeches. Furthermore, such a tool can be considered as helpful for making decisions on the words choice to deliver certain information, starting from messages that
have been stated by US Presidents during the US history. In this respect, some
final remarks on the obtained results point to interesting future research.
\newline
For what concerns the kinds of speech which would deviate from the
single exponent rank-size law patterns, one should consider that the
speeches present ``multifractal aspects''. In this respect, see e.g.
\cite{pavlov, ausloos2012a, ausloos2012b, drozdz}.
\newline
The analysis of how such speeches should be received has been
already a little bit discussed in \cite{ausloos2010punctuation,
ausloos2012b, wallot}. In this context, it would be interesting to
use readability tests, readability formulas, or readability metrics
for evaluating the readability of President speeches; this should be
(usually made) by observing punctuation, and counting syllables,
words, and sentences. One might extend such criteria looking for
``word correlations'', and in the present cases for the position of
the hapax in the speech. Some readability formulas refer to a list
of words graded for difficulty. For what we have seen, the hapaxes
are not ``difficult to understand" words (see the Section \ref{hapaxdiscussion} and Table \ref{most}).
\newline
Of course, one could also debate about the difference between
``readability" and ``understanding"; moreover reading and hearing
concern two different senses. Here we assume them to be rather
identical. 
\newline
In addition, we are able to assess a cross-document view
through low-frequency hapaxes, which may point to connection among
speeches. Such a challenging research theme might stimulate work for
future research, mainly in the linguistic arena of information science.
\newline
In the same line, the study can be extended to the case of poly-grams said only once by taking into full consideration the presence of some correlations among them, to remove crucial points as those raised in Subsection \ref{poly-grams}.
\newline
From the theoretical perspective, a promising research direction consists in formalizing the generative process for the cross-texts hapaxes' rank-size behaviour. In the environment of the rank-size procedures, the assessment of the model behind the final distribution of the ranked data is
\textit{per se} of scientific relevance. We are not aware of scholars dealing with the connection between the underlying stochastic process and the ranked phenomenon. The approach for the
construction of a probabilistic model related to a rank-size law is grounded on the interpretation of the resulting rank-size distribution the outcome of a stochastic process. In a preferential
attachment context, the idea is to define a step-wise procedure in the framework of the urn and in presence of rules stating the addition of balls in the urn at every step, as in Polya's process. An
example is found in \cite{ausloos}, where a rank-size law was also discovered. The Polya urn stochastic structure procedure can be meaningful and valid here -- under
the obvious requirement that the asymptotic distribution of the stochastic process represents a statistically significant approximation of the rank-size law.
\newline
{Finally, notice that the suggestions and the conjectures presented in Subsections \ref{hapaxexplanation} and \ref{conj} can be seen as a relevant ground for proceeding in the analysis of the hapaxes in the considered corpus of the US Presidents' speeches. }
\newline
This said, we point out that the analysis of the hapaxes here carried out provides a wide number of research questions, and that is exactly the philosophical target of our research activity.

%

\section*{Acknowledgements}
We are thankful to the Editor and the anonymous reviewers for the valuable suggestions provided during the peer review process.


\end{document}